\documentclass[journal,review,twoside]{IEEEtran}
\usepackage{bm}
\usepackage{xspace}
\usepackage{amsbsy}
\usepackage{amssymb}
\usepackage{amsmath}
\usepackage{amsfonts}
\usepackage{euscript}

\usepackage{catchfile}
\newcommand{\getenv}[2][]{%
   \CatchFileEdef{\temp}{"|kpsewhich --var-value #2"}{\endlinechar=-1}
   \if\relax\detokenize{#1}\relax\temp\else\let#1\temp\fi}

\def \bb#1{\ensuremath{{\bf #1}\xspace}}

\newcommand{\bdm}{\begin{displaymath}}
\newcommand{\edm}{\end{displaymath}}

\usepackage{graphicx} 
\usepackage{amsmath} 
\usepackage{amssymb}  
\usepackage{float}
\usepackage{dblfloatfix}
\usepackage[section]{placeins}

\usepackage{graphics} 
\usepackage{epsfig} 
\usepackage{mathptmx} 
\usepackage{times} 
\usepackage{amsmath} 
\usepackage{amssymb}
\usepackage{pifont}
\newcommand{\cmark}{\ding{51}}%
\newcommand{\xmark}{\ding{55}}%
\usepackage{algorithm}
\usepackage{algpseudocode}
\usepackage{amsmath}
\usepackage{caption}

\usepackage{color}

\usepackage{amsmath, amssymb}
\usepackage{makecell}
\usepackage{multirow}

\usepackage{mathrsfs} 
\usepackage{comment}

\usepackage{graphicx}
\usepackage{subfigure} 

\usepackage{ulem}

\usepackage{color}
\definecolor{common}{RGB}{219, 48, 122}
\definecolor{r1}{RGB}{219, 48, 122}
\definecolor{r2}{RGB}{219, 48, 122}
\definecolor{r3}{RGB}{219, 48, 122}
\definecolor{r4}{RGB}{219, 48, 122}

\usepackage{makecell}

\usepackage{graphicx}
\usepackage{float}

\newfloat{figtab}{htb}{fgtb}
\makeatletter
  \newcommand\figcaption{\def\@captype{figure}\caption}
  \newcommand\tabcaption{\def\@captype{table}\caption}
\makeatother

\usepackage{tablefootnote}

\usepackage{framed}

\usepackage{hyperref}

\ifCLASSINFOpdf
\else
\fi
\begin{document}
%
\title{Control Map Distribution using Map Query Bank for Online Map Generation}

%
%
%
\author{
Ziming Liu$^{1}$, Leichen Wang$^{1}$, Ge Yang$^{1,2}$, Xinrun Li$^{1,3}$, Xingtao Hu$^{1}$, Hao Sun$^{1}$, Guangyu Gao$^{4}$

\thanks{All authors are with $^{1}$ AID-OMG team, Bosch Research, Shanghai, China.  \\$^2$ University of Stuttgart, Stuttgart, German. $^3$ Newcastle University, Newcastle upon Tyne, England. $^{4}$ Beijing Institute of Technology, Beijing, China \\Email: \{firstname\}.\{lastname\}@cn.bosch.com}
}
%
%

\markboth{ Mar~2025}%
{LIU \MakeLowercase{\textit{et al.}}: title}
%



\maketitle

\begin{abstract}
Reliable autonomous driving systems require high-definition (HD) map that contains detailed map information for planning and navigation. However, pre-build HD map requires a large cost. Visual-based Online Map Generation (OMG) has become an alternative low-cost solution to build a local HD map. Query-based BEV Transformer has been a base model for this task. This model learns HD map predictions from an initial map queries distribution which is obtained by offline optimization on training set. Besides the quality of BEV feature, the performance of this model also highly relies on the capacity of initial map query distribution. However, this distribution is limited because the limited query number. To make map predictions optimal on each test sample, it is essential to generate a suitable initial distribution for each specific scenario. This paper proposes to decompose the whole HD map distribution into a set of point representations, namely map query bank (MQBank). To build specific map query initial distributions of different scenarios, low-cost standard definition map (SD map) data is introduced as a kind of prior knowledge. Moreover, each layer of map decoder network learns instance-level map query features, which will lose detailed information of each point. However, BEV feature map is a point-level dense feature. It is important to keep point-level information in map queries when interacting with BEV feature map. This can also be solved with map query bank method. Final experiments show a new insight on SD map prior and a new record on OpenLaneV2 benchmark with 40.5\%, 45.7\% mAP on vehicle lane and pedestrian area.

\end{abstract}


\ifCLASSOPTIONpeerreview
\begin{center} \bfseries EDICS Category: 3-BBND \end{center}
\fi
%
\IEEEpeerreviewmaketitle

\section{ Introduction  }
\label{sec:intro}
%
%
%




\IEEEPARstart{A}{utonomous} driving relies heavily on high-definition maps (HD Maps) \cite{HDMapoverview-JournalNavigation-2020} for navigation, decision-making, and path planning. An HD Map is an accurate and reliable representation of the environment, typically including details such as roads, lanes, road markers, traffic lights, and barriers. To create HD Maps, various multi-sensor technologies—such as LiDAR, radar, surrounding cameras, and GPS—are utilized. However, maintaining up-to-date HD maps in this way is expensive, and costs are often impossible for covering every city.

To address this issue, visual-based online map generation (OMG) methods attract more and more attention recently \cite{mapTR-2022,LaneSegNet-ICLR-2024}. OMG methods can obtain local HD map by visual perception networks. The most recent state-of-the-art networks are based on bird eye view (BEV) Transformer network \cite{bevformer-ECCV-2022}. The local HD map is learned as a rasterized BEV feature map, then map features are learned from this BEV feature map. Predicted HD map elements usually contain vehicle lanes and pedestrian crossing area \cite{LaneSegNet-ICLR-2024}. 

The performance of these query-based BEV Transformer networks mostly depend on two sides: initial map queries and  BEV feature map. 




Following the knowledge from object detection Transformer \cite{DETR-ECCV-2020}, previous OMG models all learn an initial map query embedding on the training dataset. This kind of initialization is the distribution of training samples instead of test samples \cite{LaneSegNet-ICLR-2024, mapTR-2022}. For OMG task, the out of distribution problem is more serious than object detection because roads structure are more stochastic. For a new scenario, it is hard to obtain optimal map query feature with this initial distribution. Although PriorMapNet \cite{PriorMapNet-arxiv-2024} suggest that a clustered training set distribution can improve the test set results. The gap between training set and test set always exists. Topo2D \cite{Topo2D-arxiv-2024} uses the features of 2D lane detection to initialize the map distribution of 3D lane detection. However, 2D lane detection model suffers the same problem.

Moreover, map queries learn HD map information from BEV feature map by cross-attention module. This BEV feature map is a point-level dense feature. To save computation cost and memory, map query feature is the representation of polyline instead of point in map decoder network \cite{mapTR-2022, LaneSegNet-ICLR-2024}. The point-level information of map query feature is lost in map decoder network. This makes instance-level map queries difficult to learn fine-grained HD map information from point-level BEV feature map in cross-attention module \cite{mapQR-ECCV-2024}.


For object detection tasks, the object distributions are not available. However, standard definition map (SD map) is a kind of low-cost, reliable and fresh data which can become the prior knowledge to initialize the HD map query distribution. To make the initial map query distribution suitable for each scenario, there is a point-level feature bank proposed to learn both geometry and semantic feature embedding, namely Map Query Bank (MQBank). The initial map queries of each scenario can be obtained with the SD map polyline coordinates and map query bank.

To learn better map queries with BEV feature map, MQBank is also used to provide fine-grained point features for instance-level map queries. The internal HD map predictions of map decoder network replace SD map polyline coordinates to generate point-level map queries from MQBank.


In practice, open-source SD map datasets still have a margin with commercial SD map: (i) miss-alignment with the ground truth HD map annotations; (ii) semantic information (e.g. road type, lane number) of SD map is incomplete or incorrect \cite{OpenLaneV2-NeurIPS-2023}.  Until now, there is no high-quality SD map benchmark. This work developes a package of SD map processing tools to manually check and correct SD map data, and provides a new SD map dataset based on OpenLaneV2 \cite{OpenLaneV2-NeurIPS-2023}. 

Finally, the contributions of this paper are summarized as follows. 
\begin{itemize}
    \item A new map query bank method is proposed to generate suitable map initial distributions of each scenario, which helps easier to find optimal HD map predictions for a new sample. 
    \item Map query bank is used to learn fine-grained point features for map queries, which can align the feature scale between map query features and BEV feature map. 
    \item This paper explores and solves the main problems existing in open-source SD map. A new data process toolchain and extended SD map dataset is proposed at link \footnote{https://github.com/LaoWangBosch/Map\_Query\_Bank}. 
\end{itemize}

This paper is organized as follows: Sec. \ref{sec:intro} introduces the motivation and contributions. Sec.\ref{sec:relatedworks} introduces related works of OMG. Sec. \ref{sec:method} describes the details of proposed methods.  Sec. \ref{sec:exp} provides abundant experimental results. Finally, Sec. \ref{sec:conclusion} gives a conclusion and a direction for future work.

\section{Related work}
\label{sec:relatedworks}
\subsection{BEV network}
BEV network is a foundational network for various autonomous driving tasks, such as 3D object detection \cite{bevformer-ECCV-2022,bevformerv2-CVPR-2023,LSS-eccv-2020}, online map generation \cite{mapTR-2022, LaneSegNet-ICLR-2024}. BEV network is composed with 2D image encoder, BEV feature map encoder, and decoder network. There have been two kinds of different architectures to obtain BEV feature map from visual features. One is based on attention operation. BEV feature map encoder is based on multi-layer self-attention and cross-attention with visual features \cite{bevformerv2-CVPR-2023}. Another is based on 2D-3D projection with depth and light-weight CNN encoding on the BEV feature map \cite{LSS-eccv-2020}. This work focuses on improving the map decoder network, which is compiled with most BEV networks. To improve the capability of map decoder network, there are also some explorations \cite{2denhancing3d-arxiv-2024,streamPER-ICCV-2023}. These methods either use 2D features to initialize 3D query features \cite{2denhancing3d-arxiv-2024} or use temporal features to help current queries \cite{streamPER-ICCV-2023}. 


\subsection{Online map generation}

There have been several successful end-to-end Transformer-based online map generation networks, such as MapTR \cite{mapTR-2022,maptrv2-arxiv-2023}, LaneSegNet \cite{LaneSegNet-ICLR-2024}.
MapTR \cite{mapTR-2022} proposes to use vectorized lane line to replace rasterized map representation. Specifically, an equal points matching method is proposed to optimize the predicted lane line using ground truth. LaneSegNet \cite{LaneSegNet-ICLR-2024} further proposes to use centerline-centric lane segment to model the HD map. In this way, the topology of the map can be easier to represent. 

Besides the network structure of OMG model, the map query distribution also contribute a large part for the final prediction performance. Previous methods learns a sparse and limited map query distribution on training data domain \cite{mapTR-2022,LaneSegNet-ICLR-2024,mapQR-ECCV-2024}. Recently, some works \cite{PriorMapNet-arxiv-2024} have suggested that the initial map query distribution plays a key role for final map predictions. This work improves OMG model from the perspective of query distribution instead of network structure.

\subsection{SD map prior}

Previous SD map prior-related works use deep nueral networks to encoding SD map. There have been two types of SD map encoding networks. Some of them encode vectorized SD map road lines \cite{SMERF-ICRA-2024,PriorMapNet-arxiv-2024}. These methods project vectorized SD map polyline features into rasterized BEV feature. In addition, there are also models that use convolution networks to encode rasterized SD map (regarded as an image) \cite{Pmapnet-RAL-2024,HRmapnet-ECCV-2024}. The rasterized SD map feature can be summed with rasterized BEV feature.  These methods learns a map encoder network to obtain SD map features, which can be fused into BEV feature or map query feature. Instead of being encoded into SD map features, this work first uses SD map as a prior knowledge to guide HD map decoder.

\section{Proposed method}
\label{sec:method}

\subsection{Review query-based BEV network}
\label{sec:reviewOMGmodel}
\subsubsection{Model structure}

The most popular query-based BEV network for OMG task are based on BEV Transformer \cite{bevformerv2-CVPR-2023,bevformer-ECCV-2022} architecture. Firstly, Image features are extracted with a shared Convolution neural network. $\bb{V}_{bev}$ is the query in this BEV Transformer encoder. BEV Transformer encoder learns a Bird-eye view (BEV) feature $\bb{V}_{bev}$ from image features $\bb{I}_{i}$  using self-attention $\Theta_{SA}$ and cross-attention $\Theta_{CA}$ for multi-layers.  
\begin{equation}
\begin{aligned}
    \bb{V}_{bev, k} = \Theta_{CA,k}\left( \Theta_{SA,k}(\bb{V}_{bev, k-1}), [\bb{I}_{i}], [\bb{I}_{i}]\right) 
 \\ i\in[0,N_{cam}], k\in[1,K]
\end{aligned}
\end{equation}
where $N_{cam}$ is the number of cameras, $\bb{V}_{bev} = \{\bb{v}_{u, v},...\}$ is a spatial matrix of point-level vector $\bb{v}_{u, v}$, $u,v$ are in the range of BEV matrix, $k$ is the network layer index.

In map decoder network, map queries $\bb{Q}$ learn relations with each other using self-attention $\Theta_{SA}$ and then learn HD map information from BEV features using cross-attention $\Theta_{CA}$ for multi-layers. 
\begin{equation}
    \bb{Q}_{l} = \Theta_{CA,l}(\Theta_{SA,l}(\bb{Q}_{l-1}), \bb{V}_{bev}, \bb{V}_{bev}), l \in [1,L]
\end{equation}
where $l$ is the map decoder layer index.

Initial map queries $\bb{Q}_{0}$ are learned feature embedding in training data. Finally,  each layer of map decoder will predict HD map elements using multi-layer perceptron (MLP) network, including lane and pedestrian area and topology relations between each element. Prediction results contain both semantic information (instance type and laneline type) and geometry information (centerline and left/right lane line). 


\begin{figure*}
    \centering
    \includegraphics[width=1\linewidth]{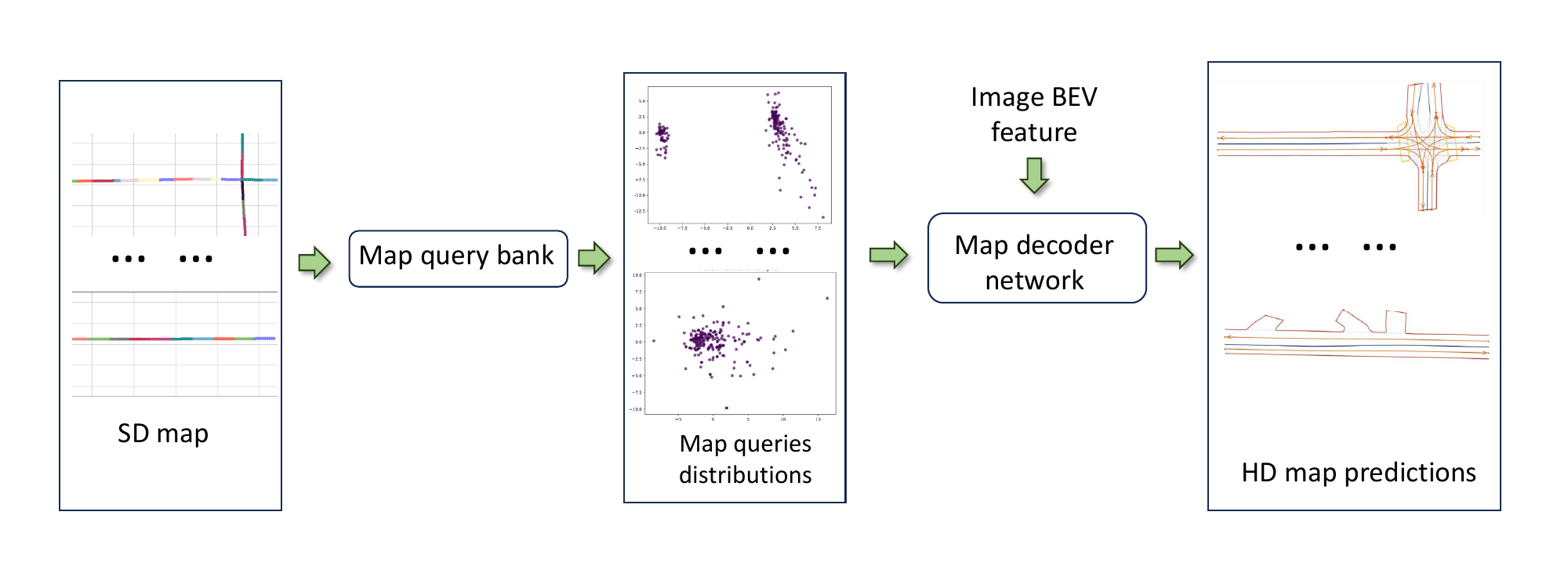}
    \caption{Online map generation model using map query bank.}
    \label{fig:model_structure0330}
\end{figure*}

\subsubsection{Optimization}
Overall, the same loss functions as \cite{LaneSegNet-ICLR-2024} are used. The predictions include both classification and position results. To optimize this model, L1 loss is used to regress geometry coordinates of all  polylines. For the classification, there are two branches. Focal loss is usually used for map element type (lane, pedestrain area, and background) classification to solve sample number unbalance. Cross-entropy loss is used for laneline type classification. In addition of the losses on vectorized map predictions, there is usually another mask segmentation branch using Cross-entropy and Dice loss to constrain this model.

\subsection{Overall structure of proposed model}

In this part, we describe the overall structure of the proposed map prior-based OMG model, as shown in Fig. \ref{fig:model_structure0330}. Firstly, this model obtains an image BEV feature same as the baseline BEV Transformer as described in Sec. \ref{sec:reviewOMGmodel}. Then, there is a new  map prior-based map decoder module. Different from the previous, initial distributions are generated from map query bank according to corresponding local SD map of different scenarios. More details are shown as follows. 




\subsection{Definition of Map query bank (MQBank)}
\label{sec:definitionmqbank}
This part introduces the definition of \textit{map query bank} $\bb{B}$. Map query contains both geometry and semantic information. 

Firstly, we describe the definition of $\bb{B}$. It is expressed as Eq. \ref{eq:geometrybank}. 

\begin{equation}
    \bb{B} =  \begin{bmatrix}
\bb{q}_{1,1} & \bb{q}_{1,j} & ... & \bb{q}_{1,w} \\
... & & & ... \\ 
\bb{q}_{i,1} & \bb{q}_{i,j} & ... & \bb{q}_{i,w}\\
...  & & & ... \\
\bb{q}_{h,1} & \bb{q}_{h,j} & ... & \bb{q}_{h,w} \\
\end{bmatrix}  
\label{eq:geometrybank}
\end{equation}

where $\bb{B}$ is a vector matrix with the shape of $i\in[1,h], j\in[1,w]$, each component is a map query geometry vector $\bb{q}$. 

Map query bank $\bb{B}$ is optimized in the training stage as learnable feature embeddings. Given map data $\bb{M}$ , corresponding map queries $\bb{Q}$ will be generated from map query bank $\bb{B}$. map data $\bb{M}$ is represented as $m$ vectorized polylines. Each polyline $ \bb{M}_i =  [(x_j, y_j), ...], i\in[1,m], j\in[1,n] $, $n$ is the number of reference points of each polyline.

Assume geometry map query $\bb{Q}$ is composed with map query vector $\bb{q}$, $\bb{Q}= {\bb{q}_i, ...}, i\in[1, m]$. 
 
Each map query geometry vector $\bb{q}$ is extracted from map query bank $\bb{B}$ as Eq. \ref{eq:indexq}. 

\begin{equation}
\bb{q} = \bb{B}(\bb{p})  
\label{eq:indexq}
\end{equation}
where position index $\bb{p}$ is a list of points $ [(u_i, v_i), ...], i\in[1,n]$, $n$ is the number of reference points. 

Position indices $\bb{P} = [\bb{p}_i, ...], i\in[1,m]$ are obtained by transforming map data $\bb{M}$. Firstly, the coordinates of map data are shifted to promise $x_{min}, y_{min} = (0,0)$. Then, position indices coordinates are transformed by re-scaling and integerization, where $u = \lfloor x \cdot \lambda +0.5 \rfloor, v = \lfloor y \cdot \lambda +0.5 \rfloor $, where $\lambda$ is the scale ratio between $\bb{B}$ and $\bb{M}$.





\subsection{Map query initialization with SD map prior}

As introduced in Sec. \ref{sec:definitionmqbank}, map query $\bb{Q}$ can be obtained given SD map coordinates. With the help of map query bank method, we can obtain a set of initial map query vectors $\bb{Q}_{init}$ from query bank $\bb{B}$ using augmented SD map data $\bb{M}'$. 

This augmented SD map data $\bb{M}'$ is obtained by random shift operation. Specifically, there is a random shift on both $x$ and $y$ directions for each SD map polyline $\bb{M}_i, i\in[1,m]$. This random shift is limited in a small range, e.g. [-10, 10]. As SD map is a kind of subset of HD map, it is essential to perform this augmentation to provide enough candidate queries for HD map prediction. In addition, SD map usually has $5 \sim 10$ meters error. At the same time, the augmented SD map polylines will also inherit semantic attribution, i.e. road type, of original SD map.


\subsection{A new attention module with map query bank}
This part describes a new BEV cross-attention module which is used to replace the attention module in Sec. \ref{sec:reviewOMGmodel}. 
Assuming $\bb{q}_{ins} = \bb{Q}(k), k\in[1,m]$ is the instance query of one polyline, each instance query can decode $n_{pts}$ reference points.
\begin{equation}
    \bb{p} = \bb{f}_{MLP}(\bb{q}_{ins}) 
\end{equation}
where $\bb{p}=[(x_1,y_1), (x_2,y_2), ..., (x_{n_{pts}},y_{n_{pts}}) ]$. $ \bb{f}_{MLP}$ is a reference points decoding network.  

In previous methods, instance query $\bb{q}_{ins} $ is directly updated to $\bb{q}'_{ins}$ as follows. 
\begin{equation}
    \bb{q}'_{ins} = softmax( \frac{\bb{q}_{ins} \cdot \bb{V}_{bev}(\bb{p}(i)) }{\sqrt{d}}) \cdot \bb{V}_{bev}(\bb{p}(i))
\end{equation}
where $d$ is a rescale factor.

In contrast, the proposed MQBank attention update map query in point-level.  Each point query $\bb{q}_{pts}$ can be generated from MQBank $\bb{B}$ with $(x, y)$ as Eq. \ref{eq:ptsquery}.

\begin{equation}
    \bb{q}_{pts,i} = \bb{B}(\bb{p}(i)), i\in[1,n_{pts}]
\label{eq:ptsquery}
\end{equation}

Then, each point query $\bb{q}_{pts}$ is updated with corresponding instance query $\bb{q}_{ins}$ as follows. 
\begin{equation}
    \bb{q}'_{pts,i} = \bb{q}_{pts,i} + \bb{q}_{ins}, i\in[1,n_{pts}]
\end{equation}

Cross-attention between point query $\bb{q}'_{pts,i}$ and  BEV feature map $\bb{V}_{bev}$ is performed as follows. 

\begin{equation}
\begin{aligned}
 & \bb{q}'_{pts, i} = softmax(\frac{\bb{q}'_{pts,i} \cdot \bb{V}_{bev}(\bb{p}(i))}{\sqrt{d}}) \cdot \bb{V}_{bev}(\bb{p}(i))
\end{aligned}
\end{equation}
This cross-attention is performed in point-level instead of instance-level \cite{LaneSegNet-ICLR-2024,SMERF-ICRA-2024}. 

After fusing with BEV feature, the point queries $\bb{q}'_{pts}$ of the same line instance will be concatenated and then fused to a line instance query $\bb{q}'_{ins}$. 

\begin{equation}
\begin{aligned}
    &\bb{q}''_{pts} = concat([\bb{q}'_{pts,i}, ...]), i\in[1,n_{pts}] \\ 
    &\bb{q}'_{ins} = \bb{f}_{MLP}(\bb{q}''_{pts})
\end{aligned}
\end{equation}
where $\bb{f}_{MLP}$ is a MLP network to align channel dimension.


\section{Experiments}
\label{sec:exp}

\subsection{Rethink SD map data}
\label{sec:sdmapdata}
\begin{figure*}[ht]
\centering
\includegraphics[width=0.9\textwidth]{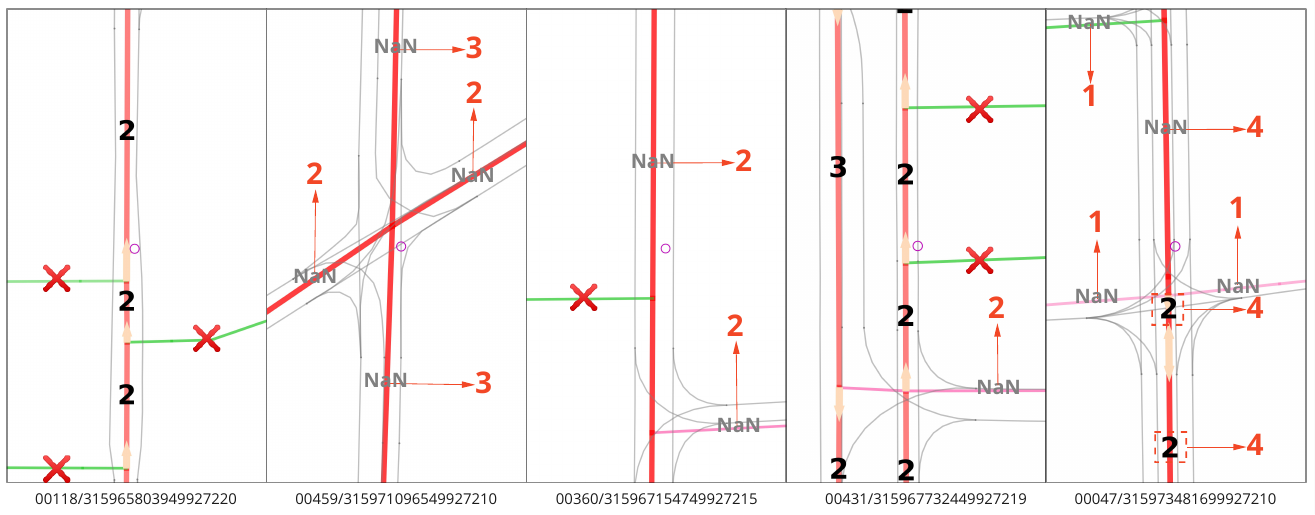}
  \caption{SD map rectification on OpenlaneV2 \cite{OpenLaneV2-NeurIPS-2023}  dataset. Colorful lines are SD map routes, gray lines are HD map driving centerlines. $\xrightarrow{}$ is to modify the number of lane on this road, which is missing or incorrect. $\times$ is to remove an extra road on SD map. Frame ID is recorded below each map.  Please refer to \href{https://github.com/LaoWangBosch/Map_Query_Bank}{this link} for more details.}
  \label{fig:sdmapshow}
\end{figure*}

While prior studies have utilized SD map priors for OMG tasks, the inherent quality limitations of SD map data remain unexplored\cite{SMERF-ICRA-2024}. After human rectification on OpenLaneV2 HD map annotation and OSM SD map data \footnote{https://www.openstreetmap.org}, three main problems are found within 850 scenes \cite{OpenLaneV2-NeurIPS-2023}. These problems are shown in Fig. \ref{fig:sdmapshow}

First, there is a false negative (FN) problem existing in HD map annotation. HD map annotations do not comprehensively cover all roads. Specifically, one road maybe exists in visual images and SD map data, but HD map annotation misses this road. This misalignment will cause an error when SD map prior is introduced.

Second,  SD map data has false positives (FP) problem. Specifically, one road exists in HD map, but this road is untraceable in SD map data. This discrepancy significantly degrades the performance of online map perception task.

Finally, semantic information in SD map data also has error information, especially for road type and lane number per road. For example, lane number per road information is missed or not correct. 

To address these problems, we developed an SD map verification and correction tool chain.  This tool performs a semi-automated comparison between HD map annotation and SD map prior, allowing uses to fix SD map errors quickly. By facilitating this process, we aim to improve the accuracy and reliability of SD map data. High-quality SD map data is essential for the research of SD map prior. This new tool will be publicly available at \href{https://github.com/LaoWangBosch/OMG_SD_map_prior_distribution}{this link}.

\subsection{Details}

The results in this paper are recorded on OpenLane-V2 dataset. This dataset contains 1,000 scenes, with each scene 30 key frames at 2 Hz.  This benchmark is OpenLaneV2 subset A  which is generated from Argoverse V2 dataset and OpenStreetMap (OSM) open-sourced SD map database. Same as HD map annotation, SD map data has the BEV range of 50m in-front and behind, and 25m in-left and right. In summary, samples with SD map data include about 27,000 frames in training set, and 4,800 frames in the validation set.

Experiments are conducted on Nvidia A100/A800 GPU with total batch size 8 on OpenLanev2 dataset. These models are trained for 24 epochs with a learning rate 2e-4. A linear warm-up method is used at the first 500 iterations with warm-up ratio 0.333. The learning rate is updated with ``Cosine Annealing" policy. Final minimum learning rate is 2e-7. Optimizer is ``AdamW" with a weight decay 0.01. The optimization of image encoder network uses a $10\times$ smaller learning rate. ResNet 50 is used as image encoder in experiments for the settings of the model.

\subsection{Data quality of SD map data}
As discussed above, the importance of SD map data has been analyzed for the online map generation task. This experiment is conducted on a SD map prior-based model. Original OSM SD map and rectified SD map are used separately as a map prior. Detection mAP and topology accuracy between lanes are recorded in the Tab. \ref{tab:data_osm}. 
According to Tab. \ref{tab:data_osm}, there are significant differences with SD maps of different qualities. Because OSM is open-source data, the quality is lower than that of commercial maps. This result suggests that SD map prior to using reliable commercial maps can improve the base model more significantly. 

\begin{table}[ht!]
    \centering
    \setlength{\tabcolsep}{3.0mm}{
    \begin{tabular}{l|c|c|c|c|c}
    \hline
         SD map & Quality &  $DET_{l} $ & $DET_{a} $ & $DET_{avg} $ & $TOP_{ll}$   \\
    \hline
         Original OSM & Low  & 38.1 & 35.0 & 36.6 & 30.8 \\ 

         Rectified OSM & High & \textbf{39.6} & \textbf{39.6} & \textbf{39.6} & \textbf{32.0} \\
    \hline
    \end{tabular}}
    \caption{Results with different SD map data qualities, based on Open-source Map (OSM) data source. }
    \label{tab:data_osm}
\end{table}

\subsection{MQBank initialization}




Map query bank initialization method is compared with random query initialization \cite{LaneSegNet-ICLR-2024} and linear query initialization using a MLP network. As shown in Tab. \ref{tab:base_vs_sdmappriorbase}, results with MQBank initialization have obvious increases on each metric. For laneSeg net, there is 5.1\% average mAP increase. For SMERF model, there is 1.6\% average mAP increase and up to 5.2\% pedestrian area mAP increase. These results suggest that MQBank initialization with SD map prior is a better solution than other methods.

\begin{table}[ht!]
    \centering
    \setlength{\tabcolsep}{1.5mm}{
    \begin{tabular}{l|c|c|c|c}
    \hline
         Base model & init & $DET_{l} $ & $DET_{a} $ & $DET_{avg} $  \\
    \hline
         Laneseg \cite{LaneSegNet-ICLR-2024} & Random & 33.3\% & 32.8\% &33.1\% \\
         Laneseg \cite{LaneSegNet-ICLR-2024} & Linear Init & 34.3\% & 37.8\% & 36.1\% \\
         Laneseg \cite{LaneSegNet-ICLR-2024} &  MQ Init & \textbf{35.6}\% & \textbf{37.9}\% & \textbf{36.8}\%  \\ 
    \hline
        SMERF \cite{SMERF-ICRA-2024} & Random & \textbf{38.7}\% & 37.1\% & {37.9}\% \\
        SMERF \cite{SMERF-ICRA-2024} & Linear Init & 35.7\% & 40.1\% & 37.9\%    \\ 
        SMERF \cite{SMERF-ICRA-2024} & MQ Init & {36.7}\% & \textbf{42.3}\% & \textbf{39.5}\% \\ 
    \hline
    \end{tabular}}
    \caption{Results of map query bank initialization. }
    \label{tab:base_vs_sdmappriorbase}
\end{table}

\begin{figure*}[ht]
    \centering
    \includegraphics[width=1\linewidth]{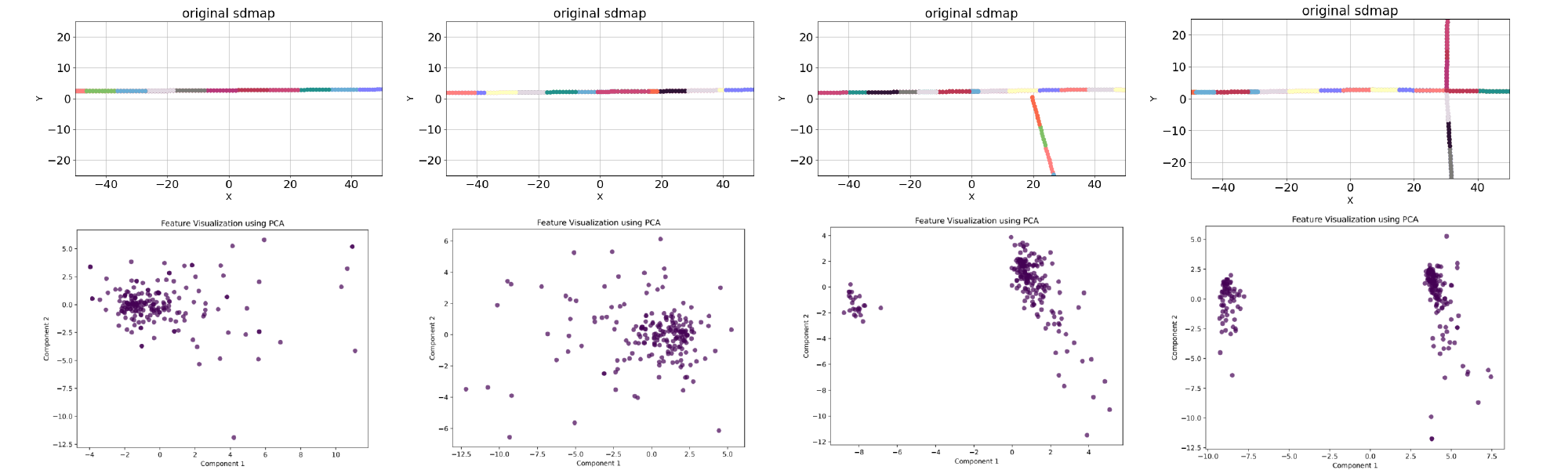}
    \includegraphics[width=0.9\linewidth]{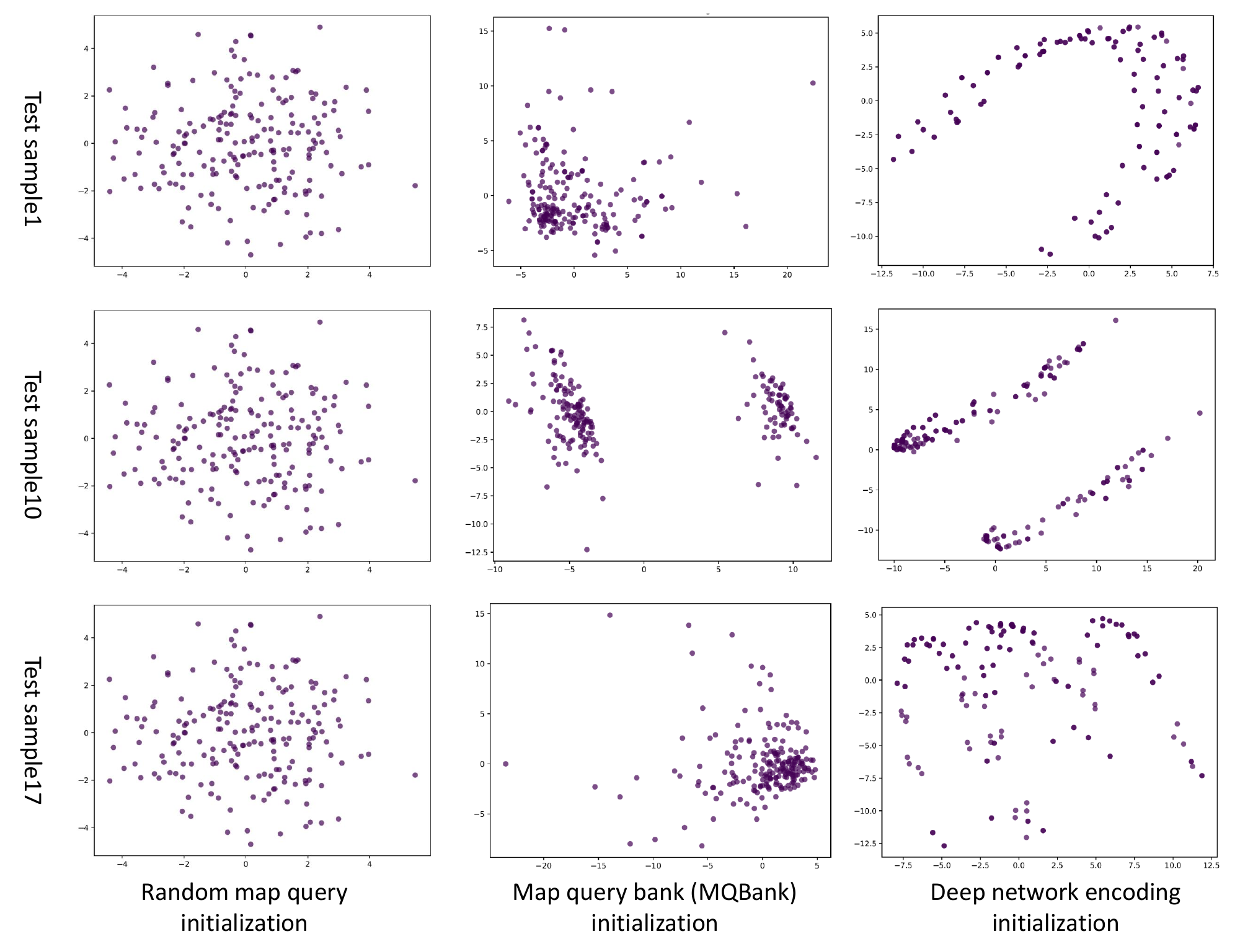}
    \caption{Comparison of distributions of different map query initialization methods. Each point is a map query feature. }
    \label{fig:pca_map_prior}
\end{figure*}

Furthermore, the visualization of different map query initializations is show as Fig. \ref{fig:pca_map_prior}. PCA is used to visualize the distributions of these map queries. Firstly, SD map and corresponding distributions from MQBank are shown. The first two rows suggest that each SD map can generate a specific distribution for each scenario. For example, the roads of x-direction and y-direction are obvious two clusters. Secondly, we compare three different map query initialization methods. The distribution of random map query initialization \cite{LaneSegNet-ICLR-2024} is uniform and same for all test samples. In contrast, MQBank initialization gives different distributions for each test samples. In addition, network linear initialization also uses SD map to generate initial map queries. However, their distributions are more discrete and unstable. Our method can learn a better feature representation for each query feature. 

\subsection{MQBank attention}

This part discusses the improvement by using map query bank attention layer in several online map generation models, including the proposed model of this paper. 
\begin{table}[ht!]
    \centering
    \setlength{\tabcolsep}{.6mm}{
    \begin{tabular}{l|c|c|c|c}
    \hline
        Model & Cross-Attention Type &  $DET_{l} $ & $DET_{a} $ & $DET_{avg}$ \\
    \hline
        \multirow{2}{*}{SMERF \cite{SMERF-ICRA-2024}} & Lane Attention \cite{LaneSegNet-ICLR-2024}& 38.7\% & 37.1\% & 37.9\% \\
          & MQBank Attention & \textbf{39.7}\% & \textbf{39.6}\% & \textbf{39.7}\% \\
         \hline
         \multirow{3}{*}{LaneSeg \cite{LaneSegNet-ICLR-2024}} &Lane Attention \cite{LaneSegNet-ICLR-2024}& 34.7\% & 39.8\% & 37.3\% \\
          &MQBank Attention &  \textbf{36.8}\% & \textbf{40.4}\% & \textbf{38.6}\% \\
          \hline
         \multirow{3}{*}{MQ Init} &Lane Attention \cite{LaneSegNet-ICLR-2024}&  36.7\% & 42.3\% &39.5\% \\
          &MQBank Attention &  \textbf{36.8}\% & \textbf{44.4}\% & \textbf{40.6}\% \\
            \hline
    \end{tabular}}
    \caption{Results of MQBank Attention. `MQ Init' uses the proposed map query bank head query initialization. }
    \label{tab:MQBank_Attention}
\end{table}

\begin{figure*}[ht!]
    \centering
    \includegraphics[width=0.32\linewidth]{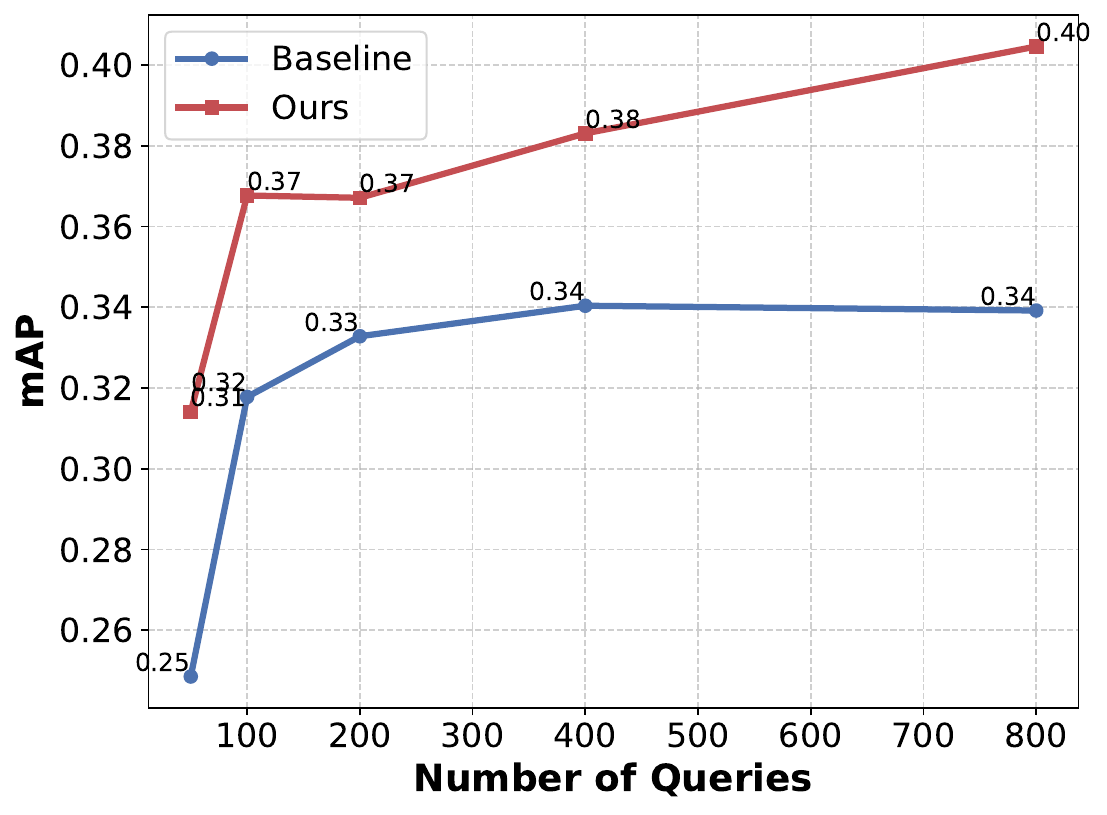}
    \includegraphics[width=0.32\linewidth]{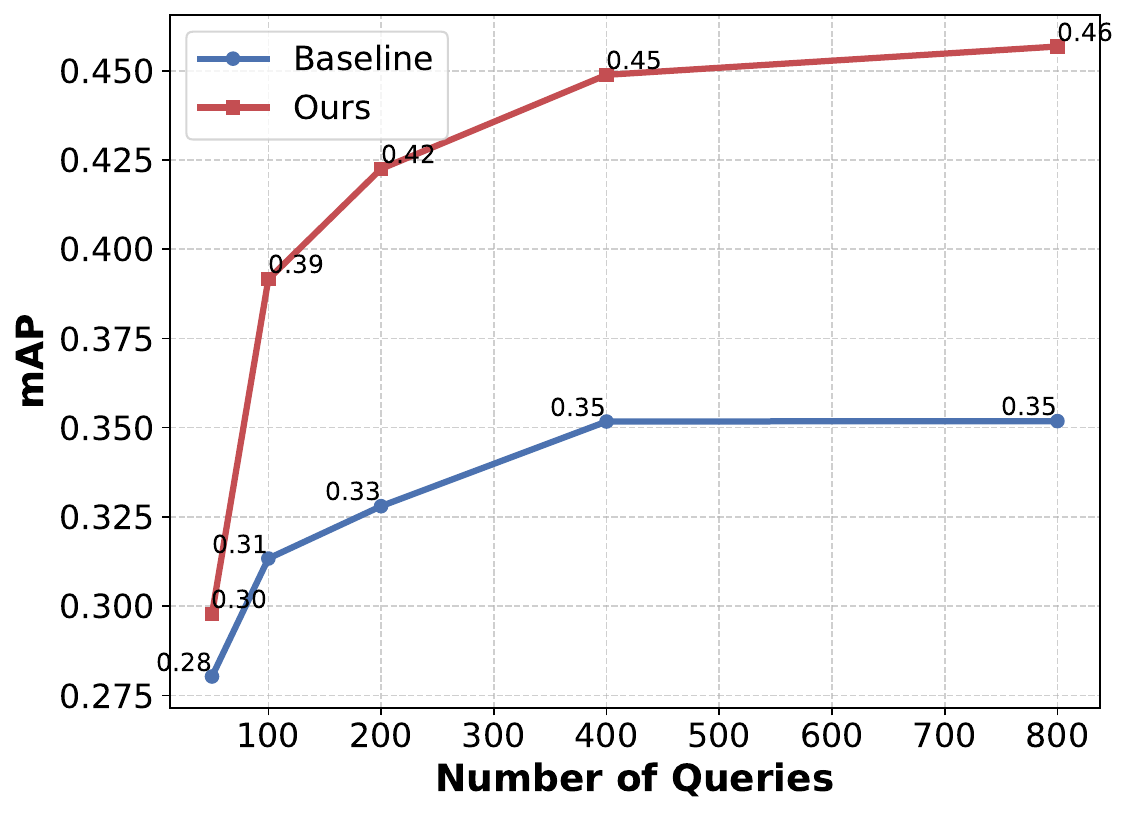}
    \includegraphics[width=0.32\linewidth]{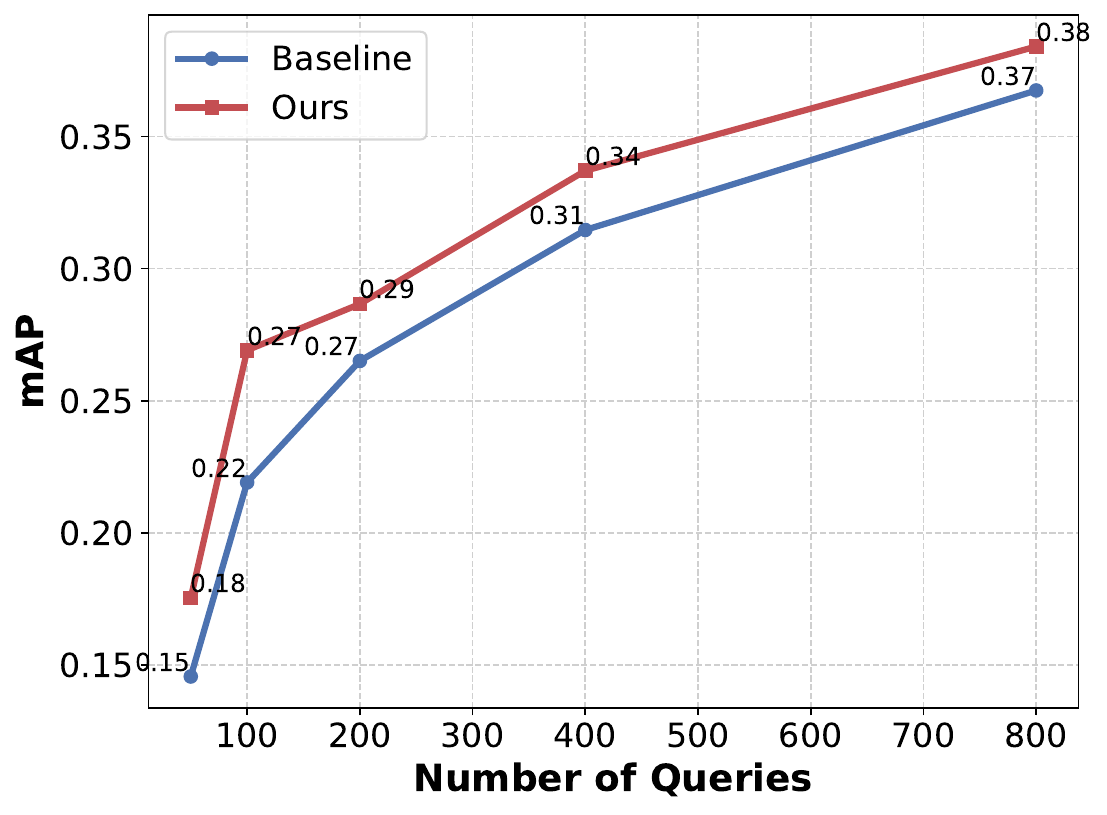}
    \caption{Comparison with or without SD map prior. Three metrics (Lane mAP, Pedestrian area mAP, Topology accuracy of lanes) are recorded, using different numbers of queries (50, 100, 200, 400, 800).}
    \label{fig:numquery}
\end{figure*}

As shown in Tab. \ref{tab:MQBank_Attention}, every metrics have significant improvements by using MQBank attention layer in decoder part of these models. Specifically, there is up to 2.5\% pedestrian area mAP improvement between Lane Attention and MQBank Attention for SMERF model \cite{SMERF-ICRA-2024}. There is up to 2.1\% lane mAP improvement for LaneSegNet model \cite{LaneSegNet-ICLR-2024}. For our model, there is also 1.1\% average detection mAP and 2.1\% pedestrian area mAP improvement.  Lane Attention \cite{LaneSegNet-ICLR-2024} is an instance-based attention module. In contrast, MQBank attention module learns map query embeddings and builds dynamic point query from map query bank. This experiment suggests the accuracy and efficiency advantage of the MQBank attention compared with the previous solutions.

\subsection{Semantic information for SD map prior-based methods}
Besides the geometry information in SD map data, semantic information also performs an important role, such as the lane number per road and road type. In this experiment, we explore different inputs with or without semantic information. Two models are used as baseline. The results are recorded in Tab. \ref{tab:semantic}. There are $>$ 2\% mAP increases for each metric. This experiment suggest the importance of SD map semantic information for online map generation.

Moreover, road type includes pedestrian area and vehicle road types. Tab. \ref{tab:semantic2} shows the results with different types of SD map roads. These results suggest that every SD map road type is useful for OMG. 

\begin{table}[ht!]
    \centering
    \setlength{\tabcolsep}{2.0mm}{
    \begin{tabular}{l|c|c|c|c}
    \hline
        Model & Road type&   $DET_{l} $ & $DET_{a} $ & $TOP_{ll}$   \\
    \hline
        \multirow{2}{*}{SMERF\cite{SMERF-ICRA-2024}} & \xmark& 34.7\% & 37.0\% & 29.2\%  \\ 
          & \cmark  & \textbf{39.0}\% & \textbf{39.4}\% & \textbf{31.6}\% \\
         \hline
         \multirow{2}{*}{Ours} & \xmark &   34.0\% & 39.7\% & 30.9\% \\ 
         & \cmark & 37.1\% & 41.3\% & 32.2\%  \\ 
            \hline
    \end{tabular}}
    \caption{Impact of semantic information (Road type) on two SD map prior-based methods. }
    \label{tab:semantic}
\end{table}

\begin{table}[ht!]
    \centering
    \setlength{\tabcolsep}{2.0mm}{
    \begin{tabular}{c|c|c|c|c}
    \hline
        \multicolumn{2}{c|}{Road type} & \multirow{2}{*}{$DET_{l} $} & \multirow{2}{*}{$DET_{a} $} & \multirow{2}{*}{$TOP_{ll}$} \\ 
        \cline{1-2}
    Pedestrian type & Vehicle type &  &  &   \\
    \hline
         \xmark & \xmark & 34.7\% & 37.0\% & 29.2\%  \\
            \xmark & \cmark & 37.9\% & 37.0\% & 30.7\%  \\ 
          \cmark  & \cmark & \textbf{39.6}\% & \textbf{39.6}\% & \textbf{32.0}\% \\
    \hline
    \end{tabular}}
    \caption{More detailed results comparison of road type.   }
    \label{tab:semantic2}
\end{table}

\subsection{Compare the results with different numbers of map queries}
With the map prior from SD map data, the proposed method can achieve better results with fewer numbers of map queries in theory. To demonstrate this assumption, Fig. \ref{fig:numquery} shows the mAP results of baseline and the proposed method on different numbers of map queries. For each metric, the baseline method requires at least double number of queries to achieve the same performance as ours. In addition, the baseline method can not achieve our average performance even with a maximum 800 queries.

\subsection{Comparison with recent online map generation methods}

The proposed method is compared with the recent online map generation models to show  the advantages of the proposed method. 
As Tab. \ref{tab:sotamethods_ol2} shows, the detection mAP of lane and pedestrian area, and lane topology accuracy have an obvious improvement compared with other methods. LaneSeg Net \cite{LaneSegNet-ICLR-2024} is the baseline for other models. Compared with this baseline, the proposed model can achieve +4.0\% lane mAP and +13.8\% pedestrian area mAP using 200 map queries. With 800 map queries, our model achieves a new state-of-the-art result on OpenLaneV2 benchmark. 

\begin{table}[ht!]
    \centering
    \setlength{\tabcolsep}{0.6mm}{
    \begin{tabular}{l|l|c|c|c|c|c|c}
    \hline
         Year & Methods& SD map & Num Q & $DET_{l} / \%$ & $DET_{a} / \%$ & $TOP_{ll} / \%$ & Latency  \\
    \hline
         2024 & LaneSeg \cite{LaneSegNet-ICLR-2024} & \xmark &200 & 32.8\% & 30.6\% & 26.3\% & 80.4ms \\ 
         2024 & MapQR \cite{mapQR-ECCV-2024} & \xmark & 200&  35.1\% & 35.7\% & 27.5\% & 82.7ms \\  
         2025 & SMART \cite{smart-arxiv-2025} & \cmark & 200 & 34.2\% & - & 16.5\% & -  \\ 
         & Ours & \cmark & 200 &  \textbf{36.8}\% & \textbf{44.4}\% & \textbf{30.3}\%& 91.3ms  \\ 
         \hline
         2024 & LaneSeg \cite{LaneSegNet-ICLR-2024} & \xmark & 800 & 33.9\% & 35.2\% & 36.7\% & 86.1ms \\  
         & Ours & \cmark & 800 &  \textbf{40.5}\% & \textbf{45.7}\% & \textbf{38.4}\% & 97.6ms \\ 
    \hline
    \end{tabular}}
    \caption{Comparison with other methods on \textit{OpenLaneV2} benchmark. Latency is measured on a server with one A800 GPU, 8 core CPU, and 50 GB memory.}
    \label{tab:sotamethods_ol2}
\end{table}






\section{Conclusion}
\label{sec:conclusion}
This paper proposes a new method, namely map query bank, to control map query distribution. Map query bank-based map decoder is different from previous map decoder networks. The distribution of map queries are explored in this paper. Based on the map query bank, we introduce a new SD map prior-based query initialization module and a new BEV cross-attention layer. Each of these new modules significantly improves the baseline model. Additionally, this paper explores the impact of SD map data on online map generation tasks. We thoroughly analyze the quality of SD map data, the semantic information it contains, etc. Due to the sparse property of SD map data, we will further investigate how to better optimize the dense map query bank in future work.


%



\section*{Acknowledgment}
This research is supported by the computational platform of Bosch Suzhou cluster.

\ifCLASSOPTIONcaptionsoff
  \newpage
\fi



%
\normalem 
\bibliographystyle{ieeetran}
\bibliography{maplessbib}

%





\end{document}